\documentclass[letterpaper, 10 pt, conference]{ieeeconf}  

\IEEEoverridecommandlockouts                              

\usepackage{graphics} 
\usepackage{epsfig} 
\usepackage{mathptmx} 
\usepackage{times} 
\usepackage{amsmath} 
\usepackage{amssymb}  
\usepackage[T1]{fontenc}
\usepackage{amsmath}
\usepackage{booktabs}
\usepackage{multicol}
\usepackage{booktabs} 
\usepackage{multirow}
\usepackage{float}
\usepackage{tabularx} 
\usepackage{amssymb} 
\usepackage{graphicx} 
\usepackage{wrapfig}  
\usepackage{tikz}     

\title{\LARGE \bf
SafeMove-RL: A Certifiable Reinforcement Learning Framework for Dynamic Motion Constraints in Trajectory Planning
}

\author{Tengfei Liu$^{1}$\textsuperscript{†},  Haoyang Zhong$^{2}$\textsuperscript{†}, Jiazheng Hu$^{1}$,  Xiaoxu Liu$^{1}$\textsuperscript{*} and Tan Zhang$^{1}$\textsuperscript{*}
\thanks{\textsuperscript{†}Equal Contribution, *Corresponding Authors.}
\thanks{This paper was supported in parts by Guangdong University Featured Innovation Program Project(2024KTSCX057), Shenzhen International Science and Technology Cooperation Project (GJHZ20220913143204009), and University and Industry Cooperation (20221061030014).}
\thanks{$^{1}$Tengfei Liu, Jiazheng Hu, Xiaoxu Liu, Tan Zhang and Jiazheng Hu are with the Sino-German College of Intelligent Manufacturing, Shenzhen Technology University, 3002 Lantian Road, Shenzhen 518118, China
        {\tt\small zhangtan@sztu.edu.cn}}%
\thanks{$^{2}$Haoyang Zhong is with Faculty of Science and Technology, University of Macau, 999078, Taipa, Macau, China
        {\tt\small }}%
}

\begin{document}

\maketitle
\thispagestyle{empty}
\pagestyle{empty}

\begin{abstract}

This study presents a dynamic safety margin-based reinforcement learning framework for local motion planning in dynamic and uncertain environments. The proposed planner integrates real-time trajectory optimization with adaptive gap analysis, enabling effective feasibility assessment under partial observability constraints. To address safety-critical computations in unknown scenarios, an enhanced online learning mechanism is introduced, which dynamically corrects spatial trajectories by forming dynamic safety margins while maintaining control invariance. Extensive evaluations, including ablation studies and comparisons with state-of-the-art algorithms, demonstrate superior success rates and computational efficiency. The framework's effectiveness is further validated on both simulated and physical robotic platforms.

\end{abstract}

\section{INTRODUCTION}

Mobile robots are increasingly being utilized in various domains, including security, exploration, and rescue, due to their superior performance and capability to significantly reduce human workload. Mobile robot often suffer from significant performance degradation in densely dynamic environments due to the inherent uncertainty of spatiotemporal constraints in dynamic free-space segmentation. While classical approaches like Dynamic Window Approach (DWA) [1] and Time Elastic Band (TEB) [2] have demonstrated effectiveness in static or semi-structured scenarios, their core optimization frameworks face critical challenges when handling time-varying obstacles with unpredictable motion patterns.  Moreover, the conventional velocity-space sampling mechanism and temporal deformation strategies frequently fail to maintain real-time performance under rapidly evolving environmental configurations, particularly in scenarios involving high-density moving obstacles with intersecting trajectories. This limitation fundamentally stems from the algorithms' insufficient capacity to simultaneously resolve spatial collision constraints and temporal feasibility in multidimensional state spaces. fundamentally stems from the algorithms' insufficient capacity to simultaneously resolve spatial collision constraints and temporal feasibility in multidimensional state spaces.

\begin{figure}[H] 
\centering 
\includegraphics[width=0.45\textwidth]{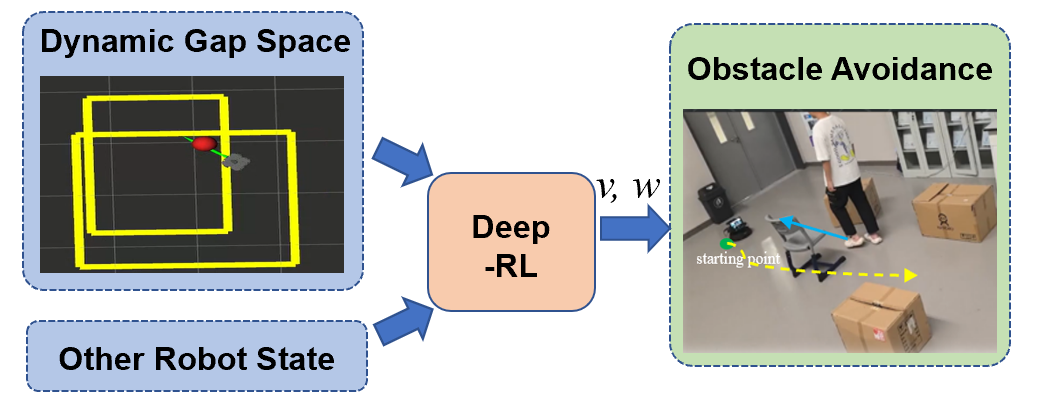} 
\caption{ Robots realize dynamic obstacle avoidance through learning of dynamic clearance and state parameters.} 
\label{Fig.1} 
\end{figure}
\vspace{-0.2cm} %

Deep reinforcement learning (DRL) has emerged as a promising solution to address computational inefficiency in dense dynamic environments. While prior studies have optimized DRL policies and network architectures[20][21], these improvements inadequately resolve the core challenge of dynamic gap navigation. Recent efforts focus on refining state-space inputs, such as integrating DWA with DRL [18] to enhance generalization, yet fail to explicitly model spatiotemporal feasibility of dynamic gaps. This deficiency becomes pronounced in scenarios requiring precise gap selection, where conventional DRL formulations struggle to balance immediate collision constraints against long-term path viability.

In this paper, we present a novel local planner that employs a model-constrained learning strategy to generate smooth, dynamic, and safe trajectories in real-time without requiring a prior map. Our approach integrates reinforcement learning techniques with model-based local planning to enable efficient online replanning. Unlike traditional methods that define the environment as a state space, our approach uses the generated local trajectories as the observation space. This new formulation allows us to impose constraints directly within the observation space, enhancing the smoothness and safety of the generated paths. Additionally, we propose an innovative reward function designed to optimize both collision avoidance and spatial feasibility. We also implement a sequential experience replay mechanism to improve training data extraction and accelerate convergence. Our method demonstrates significant improvements in responsiveness and safety in dynamic, unstructured environments. The main contributions of this work are summarized as follows: 

1) To address challenges in dynamically uncertain environments, a reinforcement learning-based adaptive safety assessment module is designed. It resolves safety planning dilemmas through online learning of probabilistic behavior patterns of dynamic obstacles, ensuring reliable collision avoidance and trajectory optimization in complex scenarios. 

2) This study proposes an innovative dynamic gap analysis framework, which establishes an efficient spatiotemporal evolution model and optimizes trajectory generation mechanisms, thereby significantly enhancing real-time performance and computational efficiency in dynamic obstacle avoidance planning.

3) Simulation and real-world validation: The improved algorithm demonstrates superior obstacle avoidance performance in complex scenarios compared to existing local obstacle avoidance methods, both in simulated environments and real-world tests.

\section{ RELATED WORK}

\subsection{Non-Learning Method for Dynamic Obstacle Avoidance}

Obstacle avoidance has always been an important research topic in path planning. Initial studies primarily focused on graph and search-based algorithms, which perform well in environments with known maps and fixed scenarios. However, to address issues where some environments are unknown or dynamically changing, researchers have developed methods such as Dynamic Window Approach (DWA) [1], Timed Elastic Band (TEB) [2], Optimal Reciprocal Collision Avoidance (ORCA) [3], and Artificial Potential Fields (AFP) [4]. These methods have demonstrated particular effectiveness in low-complexity dynamic environments. However, when the speed of obstacles is too high, the limitations of these methods significantly reduce their effectiveness in obstacle avoidance. In order to deal with complex and changeable environments, [5] proposed the proximity graph method to handle such environments, which, however, significantly increases the computational requirements. [6] proposed a novel anti-collision cone based on the perspective information of the robot and the environmental gap information, effectively enhancing the obstacle avoidance ability. [7] utilized the robot's intrinsic information and relevant motion constraints to improve its ability to extract environmental gaps. [8]combined the environmental gaps with local planning to accelerate obstacle avoidance in gap environments. [9] and [10] integrated the hierarchical planning framework with the gap-based method to improve the planning efficiency. However, [8] and [9] mainly focus on known static environments, and [10] lacks applications in real-world scenarios. In unknown dynamic environments, the gap-based method needs to balance rapid processing and obstacle avoidance requirements.

\subsection{Learning Methods for Dynamic Obstacle Avoidance }

The significant advantage of the obstacle avoidance method based on reinforcement learning is that it can effectively reduce the computational time consumption and greatly improve the computational efficiency. Existing studies [11][12][13] have used map space as the observation space and set collision and navigation as reward functions, which has markedly enhanced obstacle avoidance performance. Additionally, the performance of models has been further improved by modifying the learning network structure to a Transformer. However, these methods could have better generalization ability, making applying them in different environments challenging. To address this issue, some studies [14][15][16] have used obstacle avoidance trajectories or environmental images generated by traditional algorithms as the observation space during the learning process, achieving good obstacle avoidance results. Furthermore, studies on improving reward functions [17][18] have also demonstrated that reward designs considering angular and linear velocities can effectively improve the motion control of robots during obstacle avoidance. 

In recent years, a large number of research works have pointed out that the TD3 algorithm (Twin Delayed DDPG) [19] has demonstrated extremely outstanding potential in the obstacle avoidance tasks based on reinforcement learning. For instance, a simple network structure was proposed in [10] to improve the obstacle avoidance performance of TD3, which demonstrated the potential of using lightweight models to achieve efficient real-time decision-making. In addition, the integration of LSTM and GRU networks, as shown in [20] and [21], has been proven effective in capturing temporal dependencies in sequential data, thereby enhancing the data utilization and decision-making capabilities of the TD3 algorithm. Furthermore, the combination of TD3 with the DWA algorithm, as presented in [22], has significantly improved the robot's ability to learn and adapt to complex obstacle avoidance scenarios. This hybrid approach leverages the strengths of both algorithms to optimize the sampling interval parameters of DWA, resulting in more efficient and reliable path planning. Moreover, the integration of radar and image information with TD3, as discussed in [23], has enabled more comprehensive environment perception and efficient obstacle avoidance. Although numerous methods have been employed to improve and optimize the TD3 algorithm to varying degrees, leading to a certain enhancement in the algorithm's performance, delving deeply into the accuracy and timeliness of the TD3 algorithm in obstacle avoidance, as well as its generalization ability in complex and dynamic environments, remains a crucial challenge that urgently needs to be addressed in this field.
\section{METHOD}

\subsection{Motion Planning in Dynamic Environments} As shown in Figure 2, the navigation task utilizes a deep reinforcement learning (DRL) method, specifically the twin-delayed deep deterministic policy gradient (TD3) algorithm, to address the control problem in the continuous action space. The dual-critic network architecture effectively mitigates the overestimation bias in the function, thereby enhancing learning efficiency and model stability. In this framework, the robot does not directly use raw sensor data as input; instead, it generates an initial path by processing its own speed and environmental information. This path is then interacted with the dynamic boundary to produce corresponding path information variables, which are further processed as input to the network. Ultimately, the network outputs the required angular and linear velocities to guide the robot in performing safe and optimized motion control.

\begin{figure*}[t]
    \centering
    \includegraphics[width=\textwidth]{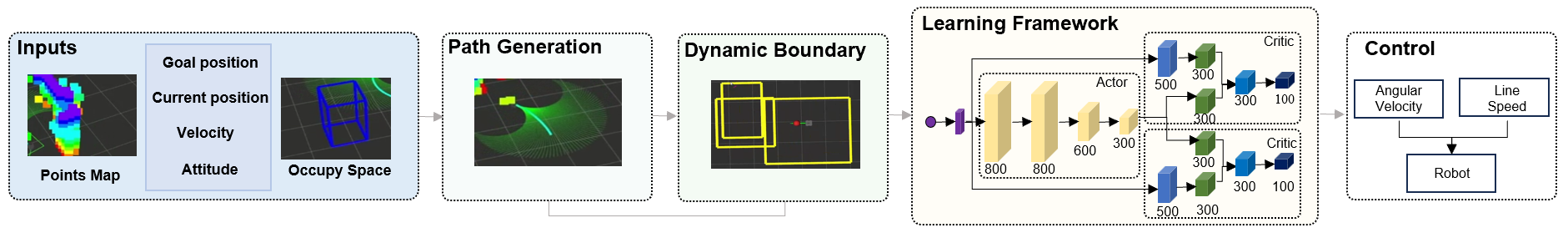}
    \caption{The overall framework of the proposed algorithm. The framework initially acquires environmental data, including point cloud occupancy information, target point direction, and corresponding velocity. This data is then used to derive DWA path information. The path information is processed to generate a dynamic boundary, which interacts with the DWA algorithm to accelerate path selection. The generated dynamic boundary and DWA path information are subsequently integrated into the deep reinforcement learning (DRL) framework to determine the angular and linear velocities required for the mobile robot.}
    \label{fig. 2}
\end{figure*}
\vspace{-0.99cm} %

\subsection{Setting Up the Observation Space} The algorithm does not directly read raw sensor data; instead, it makes decisions based on processed sensor in formation and employs complementary methods to optimize obstacle information processing. Initially, a map is generated using depth data to accurately detect surrounding dynamic and static obstacles. Concurrently, irregular obstacles are managed using the bounding box method, as proposed in [21][24], to enhance efficiency in complex environments. Path planning is then conducted in free space to create discrete paths and safe corridors. An improved DWA is employed to generate candidate paths, incorporating a novel obstacle density cost function to enhance obstacle avoidance performance, as illustrated in Fig 3. Candidate paths are selected based on obstacle count to improve the safety of path selection. The cost function of the improved DWA algorithm is denoted as  \textit{G(v, w)}:

\begin{equation}
\label{eq:cost_function}
G(v, w) = \sigma \Bigl[ 
\begin{aligned}[t]
    & \alpha \cdot \text{heading}(v, w) + \beta \cdot \text{dis}_{\text{obst}}(v, w) \\
    & + \gamma \cdot \text{vel}(v, w) + \delta \cdot \text{density}_{\text{cost}}(v, w)
\Bigr]
\end{aligned}
\end{equation}

where: $\alpha$, $\beta$, $\gamma$, $\delta$ : are the weight constants for each cost function, determining the importance of each cost term in the overall objective function. $\sigma$ is a normalization function to ensure that the value of the objective function remains within a reasonable range.
    \textit{heading}$(v, w)$: Evaluates how well a velocity pair aligns with the target direction, with higher values for better alignment.
    \textit{dis\textsubscript{obst}}$(v, w)$: Assesses the distance from obstacles, with higher values for velocity pairs farther from obstacles.
    \textit{vel}$(v, w)$: Evaluates the magnitude of the velocity, favoring higher speeds while considering other cost terms.
    \textit{distance\textsubscript{cost}}$(v, w)$: Indicates the tendency to decelerate and increase trajectory curvature as obstacle density increases.

\begin{align}
\mathit{density\_cost}(v,w) = \frac{\delta \ast N_{\mathit{obs}}}{\mathit{avg\_distance}(v,w)}
\end{align}

where: $\delta$: represents the weight coefficient used to adjust the impact of obstacle density on the total cost. 
    \textit{N\textsubscript{obs}}: represents the number of obstacles within the range of the robot's current velocity pair $(v, w)$.
    \textit{avg\_distobs}$(v, w)$: represents the average distance from the robot's trajectory, generated along the velocity pair $(v, w)$, to all obstacles.

\begin{align}
avg\_distance(v, w) = \frac{1}{M \times N} \sum_{j=1}^{M} \sum_{i=1}^{N} d_{ij}
\end{align}

where:  \ $M$: The total number of discrete time steps within a given period. \ $N$: The total number of obstacles. \ $d_{ij}$: The distance between the robot at the $i$ time step and the $j$ obstacle, defined as:

\begin{align}
d_{ij}=\sqrt{\left(x_j-x_i\right)^2+\left(y_j-y_i\right)^2}\quad
\end{align}
\begin{align}
p(t)=\left[x_p(t),y_p(t)\right]^T,t \epsilon[t_0,t_m]\quad
\end{align}

After calculating the normal vector plane of the reference point, the normal plane $\Phi$ is constructed based on the point $p(\theta)$ generated from the reference trajectory. This plane is perpendicular to the reference trajectory at the velocity direction vector $v_p^{(k)} = p(\theta^{(k)})$. The normal plane $\Phi_{k}$ is defined as:

\begin{align}
\Phi^{(k)}{:}n^{(k)}*\left[x-p\left(\theta^{(k)}\right)\right]=0
\end{align}

where: $n^{(k)}$ is the unit average vector, representing the direction of the standard plane:

\begin{align}
n^{(k)}=\frac{v_p^{(k)}}{\left\|v_p^k\right\|}
\end{align}

where: $x$ : represents any point in a two-dimensional space;

\begin{align}
x=[x,y]^{T}
\end{align}

The average plane $\Phi^{(k)}$ is defined as follows: To obtain the boundary polygon $\Omega$ on the plane $\Phi^{(k)}$, it is necessary to find all possible obstacle boundaries or predefined boundary points on the plane, denoted as $\{q_1, q_2, \ldots, q_m\}$. Then, a polygon is constructed using the convex hull of these points.

\begin{align}
\Omega^{(k)}=ConvexHull\{q_1,q_2,...,q_m\}
\end{align}

As the obstacle information changes, the robot learns the geometric position data within the convex hull polygon defined by the safe corridor.

\vspace{-0.3cm} 
\begin{figure}[H] 
\centering 
\includegraphics[width=0.4\textwidth]{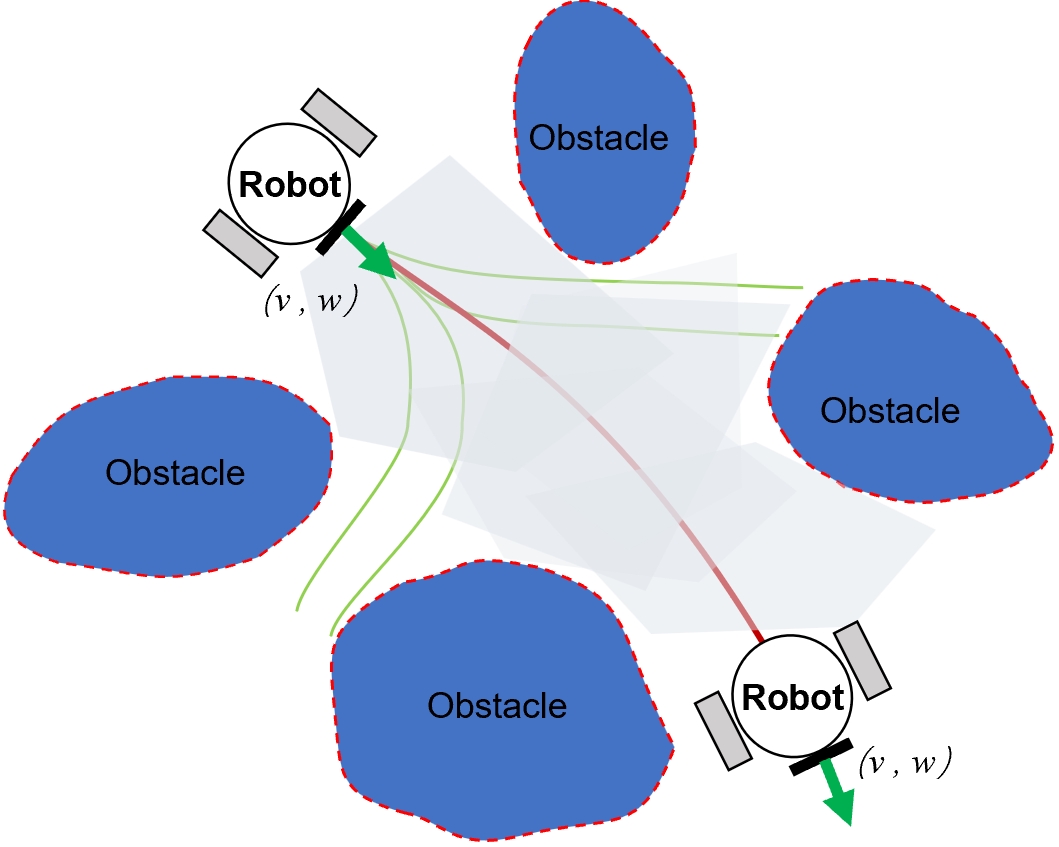} 
\caption{ illustrates the trajectory planning framework: the green line indicates the Dynamic Window Approach (DWA)-generated reference path, while the light-blue region defines the safety corridor boundaries. The red trajectory demonstrates the optimized path adhering to these constraints.} 
\label{Fig.3} 
\end{figure}
\vspace{-0.35cm} %

\subsection{Reward Function Setting} 

To achieve the core requirements of reaching the target point and avoiding obstacles while accelerating the model's training process, an appropriate total reward function $R_{total}$ needs to be designed. This function should comprehensively consider multiple factors in path planning, including goal orientation and obstacle avoidance, to balance the relationship between exploration and exploitation, thereby improving the deep reinforcement learning model's training efficiency and convergence speed. The total reward function $R_{total}$:
\begin{align}
R_{total}=R_{g}+R_{b}+R_{raw}
\end{align}
  For the target reward function $R_{g}$:
\begin{equation}
    R_g =
    \begin{cases} 
        100, & \text{if } dis(p_g , p_{g}) < 0.2m \\ 
        |v - w|, & \text{other}
    \end{cases}
    \label{eq:rg}
\end{equation}

The distance function $\text{dis}(p_r, p_g)$ is:
\begin{align}
dis(p_{r},p_{g})=\sqrt{(x_{r}-x_{g})^{2}+(y_{r}-y_{g})^{2}}\
\end{align}

Let $p_r$ and $p_g$ be the coordinates of the robot's position and the target point, respectively, with $x_r$, $x_g$, $y_r$, $y_g$ representing their corresponding x-axis and y-axis coordinates. When the robot successfully reaches the target point, a significant positive reward is given; otherwise, based on the robot's current motion state, its angular and linear velocities are adjusted to reduce the probability of deviating from the target through the design of the reward function, thereby optimizing the path planning process and improving the model's training efficiency. For the collision function $R_b$:

\begin{equation}
    R_b =
    \begin{cases} 
        -10, & \text{if } dis(p_r, p_o) \leq 0.1m \\ 
        0, & \text{others}
    \end{cases}
    \label{eq:rb}
\end{equation}


 $p_o$ denotes the position information of the obstacle. When the distance between the obstacle and the robot is less than 0.1 meters, it is considered a collision, and a negative reward of -10 is assigned to the robot. This distance threshold is set to account for the deviation between the sensor position and the robot's body, as well as the effects of point cloud data expansion, effectively representing the risk of collision. On this basis, a refined reward design guides the robot to more accurately avoid obstacles in path planning, enhancing the training effectiveness of the deep reinforcement learning model. For the orientation function $R_{yaw}$:


\begin{equation}
    R_{raw} =
    \begin{cases} 
        -10 \cdot dis(p_{start}, p_r), & \text{if } s \in G \\
        5 \cdot dis(p_{start}, p_r), & \text{if } s \in R
    \end{cases}
    \label{eq:reward}
\end{equation}

The aforementioned G and R respectively belong to the subspace within the safety corridor and the space outside the safety corridor. When the robot is oriented toward the safety corridor, its spatial perception ability of dynamic obstacles can be enhanced. In this environment, the generated path can avoid obstacles to the greatest extent, further improving the safety and efficiency of path planning.

\subsection{Policy Framework Settings}
To enhance the learning performance of the model, we aim to achieve optimal action strategy performance by further optimizing learning efficiency. We adopted the TD3 (Twin Delayed Deep Deterministic Policy Gradient) algorithm, a reinforcement learning method specifically designed for continuous action spaces. However, the traditional TD3 algorithm will affect the overall performance of the algorithm due to the randomness of the Reply buffer, and we have improved it as shown in Fig 4. 

\vspace{-0.3cm} 
\begin{figure}[H] 
\centering 
\includegraphics[width=0.45\textwidth]{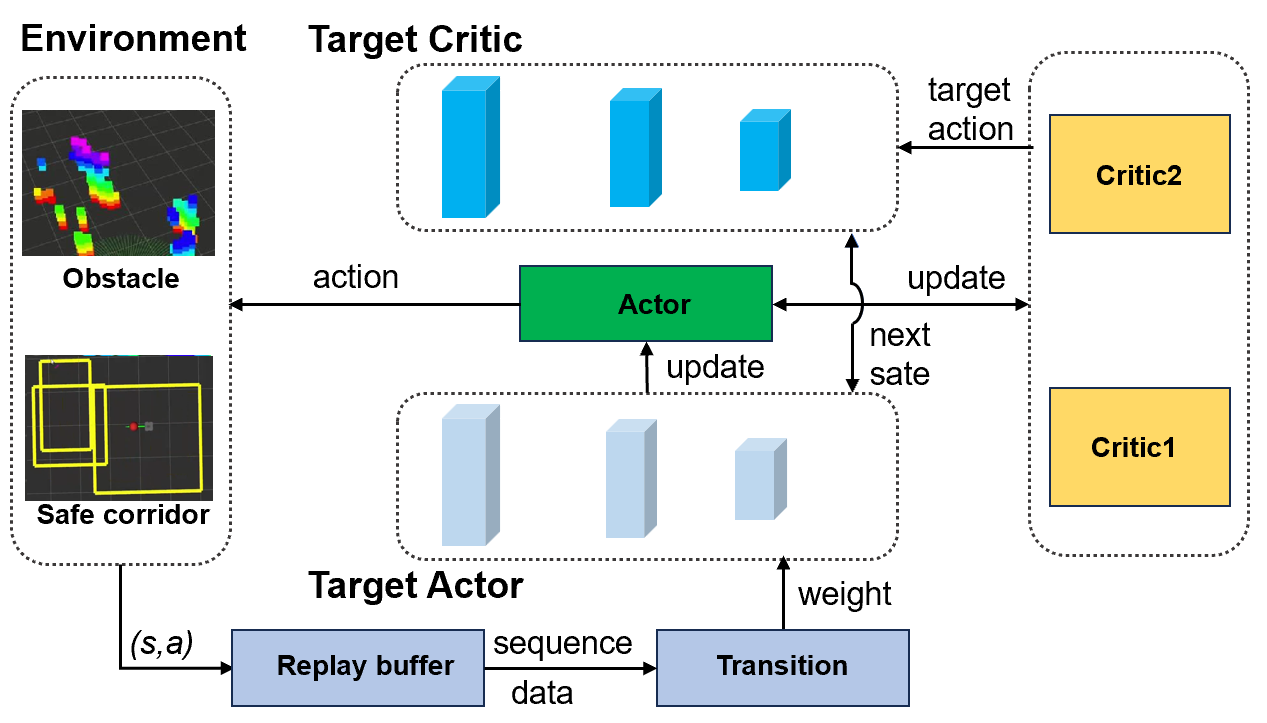} 
\caption{Policy Framework Settings.} 
\label{Fig.4} 
\end{figure}
\vspace{-0.35cm} %

In the improved model architecture, the algorithm relies on the sampled state transition pairs $(s_t, a_t, s_{t+1})$, which are stored in the replay buffer. After each action execution, the replay buffer sorts and filters the obtained reward signals (R values) to determine the maximum reward value ($R_{max}$). This process optimizes experience selection through the replay mechanism in the gray-white function to extract the optimal experience values. The selected experience values are then fed into the actor and critic networks of the TD3 algorithm, generating the following action $a_{t+1}$ through the policy network. With each action execution, the algorithm continuously updates the policy, ultimately achieving optimal path planning and strategy optimization.

\begin{table*}[ht]
\centering
\caption{Ablation experiment: Performance of different algorithms in different dynamic obstacle environments}
\label{tab:ablation_study}
\begin{tabularx}{\textwidth}{
    >{\centering\arraybackslash}c
    >{\centering\arraybackslash}c
    >{\centering\arraybackslash}X
    >{\centering\arraybackslash}c
    >{\centering\arraybackslash}c
    >{\centering\arraybackslash}c
    >{\centering\arraybackslash}c
}
\toprule
\textbf{Environment} & 
\textbf{Dynamic obstacle} & 
\textbf{Method} & 
\textbf{Success} & 
\textbf{Collision} & 
\textbf{Average length(m)} & 
\textbf{Average velocity(m/s)} \\
\midrule

\multirow{3}{*}{\centering (a)} & 
\multirow{3}{*}{\centering 8} & 
SafeMove-RL (Ours) & 100\% & 0\% & 7.25 & 0.19 \\ 
& & DRL+DWA & 95\% & 5\% & 7.16 & 0.20 \\
& & DRL & 100\% & 0\% & 7.33 & 0.17 \\
\midrule

\multirow{3}{*}{\centering (b)} & 
\multirow{3}{*}{\centering 12} & 
SafeMove-RL (Ours) & 95\% & 5\% & 7.44 & 0.18 \\ 
& & DRL+DWA & 85\% & 15\% & 7.49 & 0.18 \\
& & DRL & 90\% & 10\% & 7.38 & 0.16 \\
\midrule

\multirow{3}{*}{\centering (c)} & 
\multirow{3}{*}{\centering 16} & 
SafeMove-RL (Ours) & 85\% & 15\% & 7.63 & 0.16 \\ 
& & DRL+DWA & 70\% & 30\% & 7.60 & 0.18 \\
& & DRL & 70\% & 30\% & 7.58 & 0.13 \\
\midrule

\multirow{3}{*}{\centering (d)} & 
\multirow{3}{*}{\centering 20} & 
SafeMove-RL (Ours) & 80\% & 20\% & 7.98 & 0.17 \\ 
& & DRL+DWA & 40\% & 60\% & 7.66 & 0.18 \\
& & DRL & 60\% & 40\% & 7.48 & 0.14 \\
\bottomrule
\end{tabularx}
\end{table*}

\section{EXPERIMENTS }

To validate the effectiveness of the proposed local path planning algorithm, we conducted a series of experiments in both simulated and real-world environments. During the simulation phase, ablation and comparative experiments were performed using the Turtlebot3 differential-drive robot. The tests were carried out on a platform equipped with an Intel Core i9-13700K CPU and an NVIDIA RTX 4000 GPU. In the real-world testing phase, the algorithm was deployed on a Spark differential-drive robot, which is equipped with an Intel Core i5-8259U CPU and an integrated Intel Iris Plus Graphics 655 GPU, to evaluate its performance in practical scenarios.

\subsection{Training settings}

\begin{figure}[H] 
\centering 
\includegraphics[width=0.4\textwidth]{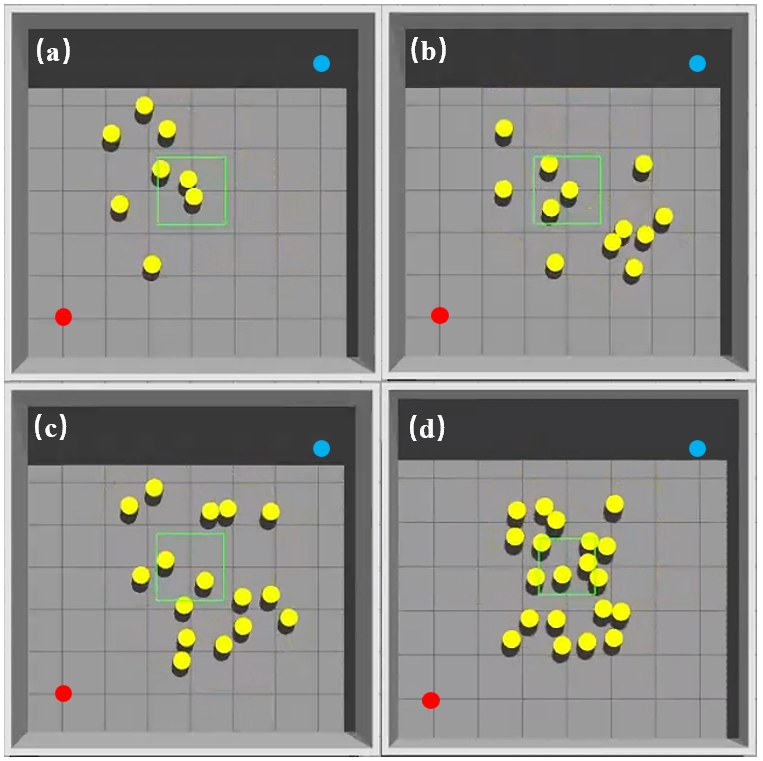} 
\caption{Four simulation environments with color-coded components: yellow cylinders denote dynamic obstacles governed by ORCA algorithm, blue spheres represent predefined target points, and red rectangular areas indicate unified initial positions. Environmental complexity increases from (a) to (d) through incremental addition of dynamic obstacles (8, 12, 16, 20), while maintaining consistent geometric constraints for target/start configurations.} 
\label{Fig.5} 
\end{figure}
\vspace{-0.35cm} %

\subsubsection{ Training Environment}

This study systematically evaluates robotic navigation algorithms by constructing four experimental scenarios with graded complexity levels in the Gazebo simulation platform (Figure \ref{Fig.5}). The test environments feature progressively complex configurations, achieved through controlled increments of dynamic obstacles (8, 12, 16, and 20 respectively). All dynamic obstacles implement Optimal Reciprocal Collision Avoidance (ORCA) algorithm\cite{c25} for motion planning, with randomized velocities uniformly distributed between $0.5$ and $1.0$ m/s. This stratified experimental framework rigorously assesses algorithm robustness against escalating dynamic disturbances while maintaining parametric consistency across trials.

\subsubsection{ Evaluation Metrics}

We evaluated the performance of our trained model in a dynamic environment by setting up a consistent starting point and four target points, each with different orientations. At each target point, we conducted ten sets of experiments. To assess the performance of the local path planner in comparison to other algorithms in terms of safety and planning efficiency, we established three key evaluation metrics:

\begin{enumerate}
    \item \textbf{Success Rate}: Indicates the ratio of the number of times the robot successfully reaches the goal point to the total number of attempts, measuring the ability of the algorithm to accomplish the task.
    
    \item \textbf{Average Trajectory Length}: The total traveling path length when the robot successfully reaches the goal point divided by the number of successes, assessing the efficiency and simplicity of path planning.
    
    \item \textbf{Average Line Speed}: The total path length divided by the total time when the robot successfully reaches the goal point, divided by the number of successes, reflecting the efficiency of the robot in moving along the path.
\end{enumerate}

\begin{table*}[ht]
\centering
\caption{Baseline comparison of different number of obstacles and density in an unknown environment}
\label{tab:baseline_centered}
\begin{tabularx}{\textwidth}{
    >{\centering\arraybackslash}c
    >{\centering\arraybackslash}c
    >{\centering\arraybackslash}X
    >{\centering\arraybackslash}c
    >{\centering\arraybackslash}c
    >{\centering\arraybackslash}c
    >{\centering\arraybackslash}c
}
\toprule
\textbf{Environment} & 
\textbf{Dynamic obstacle} & 
\textbf{Method} & 
\textbf{Success} & 
\textbf{Collision} & 
\textbf{Average length(m)} & 
\textbf{Average velocity(m/s)} \\
\midrule

\multirow{6}{*}{\centering (a)} & 
\multirow{6}{*}{\centering 8} & 
DWA [1] & 95\% & 5\% & 7.22 & 0.20 \\ 
& & TEB [2] & 90\% & 10\% & 7.34 & 0.19 \\
& & Potential Gap [8] & 100\% & 0\% & 7.47 & 0.20 \\
& & TD3-Nav [11] & 95\% & 5\% & 7.26 & 0.17 \\
& & DWA-RL [18] & 100\% & 0\% & 7.33 & 0.18 \\
& & SafeMove-RL (Ours) & 100\% & 0\% & 7.25 & 0.19 \\
\midrule

\multirow{6}{*}{\centering (b)} & 
\multirow{6}{*}{\centering 12} & 
DWA [1] & 90\% & 10\% & 7.65 & 0.20 \\ 
& & TEB [2] & 80\% & 20\% & 7.47 & 0.17 \\
& & Potential Gap [8] & 85\% & 15\% & 7.37 & 0.19 \\
& & TD3-Nav [12] & 85\% & 15\% & 7.96 & 0.20 \\
& & DWA-RL [18] & 90\% & 10\% & 8.09 & 0.19 \\
& & SafeMove-RL (Ours) & 95\% & 5\% & 7.44 & 0.18 \\
\midrule

\multirow{6}{*}{\centering (c)} & 
\multirow{6}{*}{\centering 16} & 
DWA [1] & 75\% & 25\% & 7.35 & 0.17 \\ 
& & TEB [2] & 65\% & 35\% & 7.81 & 0.18 \\
& & Potential Gap [8] & 80\% & 20\% & 8.19 & 0.17 \\
& & TD3-Nav [12] & 75\% & 25\% & 7.79 & 0.15 \\
& & DWA-RL [18] & 85\% & 15\% & 7.86 & 0.16 \\
& & SafeMove-RL (Ours) & 85\% & 15\% & 7.63 & 0.16 \\
\midrule

\multirow{6}{*}{\centering (d)} & 
\multirow{6}{*}{\centering 20} & 
DWA [1] & 65\% & 35\% & 8.16 & 0.19 \\ 
& & TEB [2] & 60\% & 40\% & 7.91 & 0.18 \\
& & Potential Gap [8] & 60\% & 40\% & 7.27 & 0.12 \\
& & TD3-Nav [12] & 65\% & 35\% & 8.27 & 0.21 \\
& & DWA-RL [18] & 70\% & 30\% & 7.62 & 0.18 \\
& & SafeMove-RL (Ours) & 80\% & 20\% & 7.98 & 0.17 \\
\bottomrule
\end{tabularx}
\end{table*}

\subsubsection{Ablation Baseline}

To systematically evaluate the contributions of individual components to the overall performance of our proposed algorithm, we conducted a series of ablation experiments. The experiments compared three variants of the algorithm, each with different combinations of components, while maintaining a consistent training framework based on deep reinforcement learning (DRL). The variants are described as follows:

\begin{enumerate}
    \item \textbf{SafeMove-RL (Ours)}: This variant integrates DRL with DWA and Safe Corridor (SC) optimization. The SC module is specifically designed to enhance path planning by incorporating safety constraints, thereby improving the robustness and efficiency of the navigation process.
    
    \item \textbf{DRL+DWA}: This variant utilizes the same raw robot data as the full algorithm but omits the SC module. Consequently, it relies solely on DRL and DWA for navigation without the additional path optimization provided by SC.
    
    \item \textbf{DRL}: This variant is further simplified by excluding the robot's path trajectory data. It relies entirely on the depth camera data as input features and focuses only on the core DRL framework for decision making.
\end{enumerate}

\subsection{ Ablation Experiment}
\subsubsection{ Training Efficiency}

To quantify the contributions of individual algorithmic components, ablation studies were conducted in Scenario (b) (12 ORCA-controlled dynamic obstacles) comparing the proposed method (\textbf{DWA-DRL-SC}) against \textbf{DWA-DRL} and \textbf{DRL} baselines.

\begin{figure}[H] 
\centering 
\includegraphics[width=0.4\textwidth]{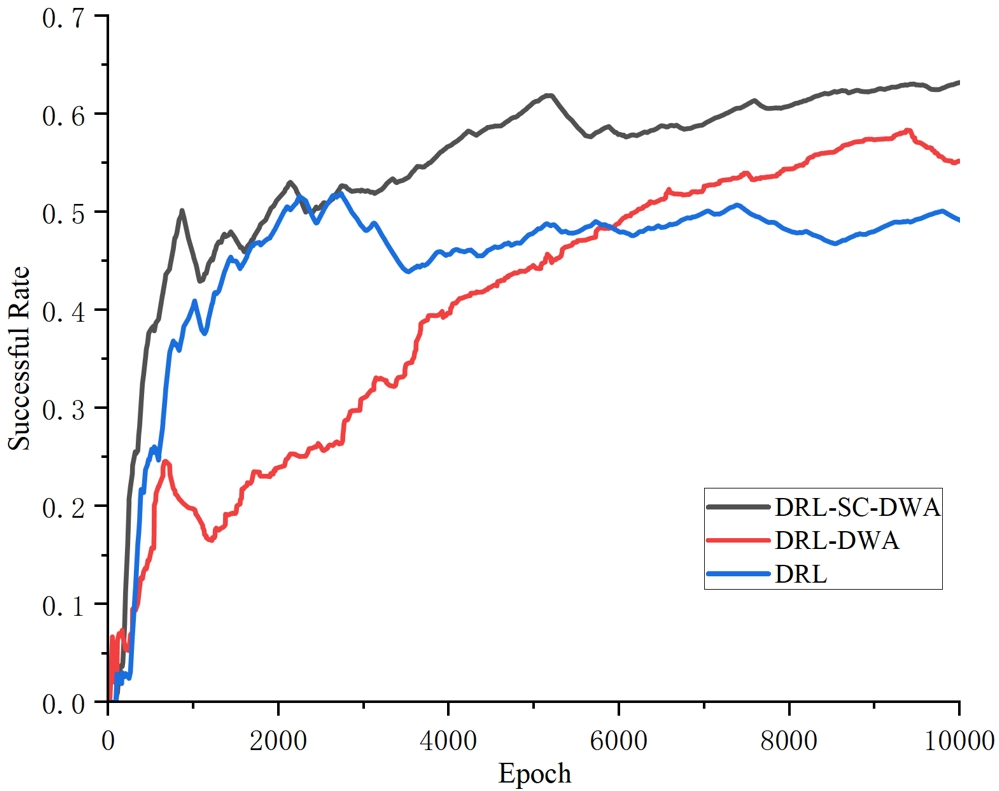} 
\caption{The convergence performance of several ablated algorithms. The red line represents the algorithm presented in this paper, the blue line represents DRL-DWA, and the green line represents the DRL algorithm.} 
\label{fig: 6} 
\end{figure}

As illustrated in Fig.~\ref{fig: 6}, our approach demonstrates superior convergence speed and final obstacle avoidance success rate. These results quantitatively validate the efficacy of the dynamic reward mechanism. Through comparative experiments, the results demonstrate distinct performance characteristics among the evaluated algorithms (Fig.~\ref{fig: 6}).

In 12 dynamic obstacle scenarios, the DWA-DRL algorithm exhibited a significantly lower obstacle avoidance success rate ($\approx 0.485$) compared to the pure DRL baseline during the initial $5{,}800$ training episodes. However, as training progressed to $10{,}000$ episodes, its success rate displayed a nonlinear increasing trend, ultimately stabilizing at $\approx 0.558$ -- a $7.2\%$ improvement over the DRL baseline. This evolution suggests that the DWA module enhances agent adaptability in complex dynamic environments through its delayed reward mechanism, enabling progressive optimization of path planning strategies.

Notably, the DWA-DRL-SC variant achieved superior learning efficiency, attaining a final success rate $8.6\%$ higher than standard DWA-DRL and $13.1\%$ above the DRL baseline. This improvement primarily stems from the algorithm's optimized exploration strategy, which effectively mitigates the initial training inefficiency caused by excessive action space exploration in conventional DWA-DRL implementations.

\vspace{-0.35cm} %

\begin{figure*}[t]
    \centering
    \includegraphics[width=\textwidth]{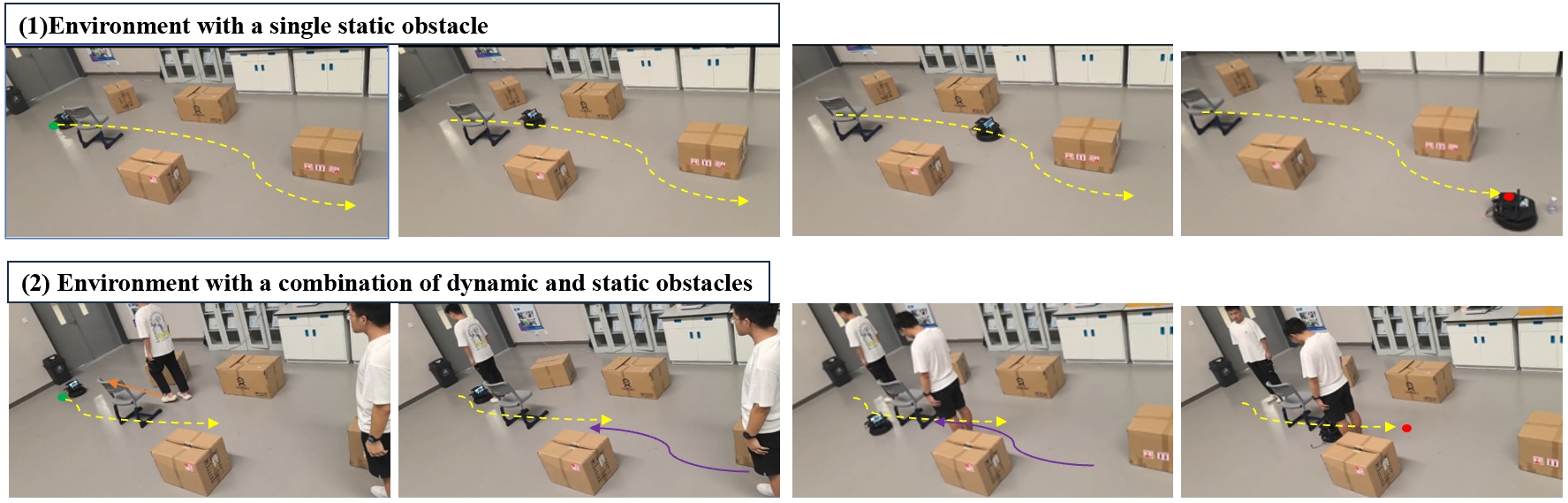}
    \caption{The obstacle avoidance performance of the robot was evaluated in both static and dynamic environments, as illustrated in Figs. (1) and (2). In the static environment (1), the robot successfully navigated around stationary obstacles, with its trajectory depicted by the yellow curve. In the dynamic environment (2), the robot effectively avoided two moving obstacles, as evidenced by its motion trajectory (yellow curve) and the corresponding motion vectors of the dynamic obstacles (blue and purple arrows).respectively. This performance discrepancy accentuates the critical role of our optimized collision prediction module in addressing dense obstacle configurations through enhanced decision-making robustness.}
    \label{fig. 7}
\end{figure*}

\subsubsection{  Ablation Simulation Performance}

The ablation study results illustrated in Fig.~\ref{fig: 6} reveal systematic performance comparisons through 10 repeated trials per experimental scenario (see Table~\ref{tab:baseline_centered} for aggregated metrics). While maintaining comparable trajectory efficiency (average speed: $0.17 \, \text{m/s}$ vs. $0.14\text{--}0.20 \, \text{m/s}$; path distance: $7.62 \, \text{m}$ vs. $7.16\text{--}7.98 \, \text{m}$) with baseline methods, our algorithm demonstrates significant advantage in navigation success rates. Particularly in high-density dynamic environments, it achieves superior obstacle avoidance performance -- outperforming counterparts by $15\%$ and $20\%$ success rate margins in scenarios containing 16 and 20 dynamic obstacles, respectively.

\subsection{Comparative Experiments}
To evaluate the performance of the proposed algorithm in localized obstacle avoidance, we conducted a comparative study against four state-of-the-art localized obstacle avoidance algorithms, including DWA\cite{c1}, TEB\cite{c2}, TD3-Navigation\cite{c11}, Potential Gap\cite{c8}, and DWA-RL\cite{c18}. The evaluation was performed across four environments with varying levels of complexity (environments a, b, c, and d) to assess three key metrics: success rate, average trajectory length, and average time.

The experimental results demonstrate that the proposed algorithm (SafeMove) exhibits superior performance in dynamic obstacle-dense environments compared to baseline methods, with its advantages becoming more pronounced as the number of dynamic obstacles increases. In scenario (a) with 8 obstacles, SafeMove achieves a $100\%$ success rate and $0\%$ collision rate, matching the best-performing method (Potential Gap). However, as obstacle density escalates, SafeMove consistently outperforms competitors. For instance, in scenario (d) with 20 obstacles, SafeMove maintains an $80\%$ success rate, significantly exceeding DWA ($65\%$), TEB ($60\%$), and Potential Gap ($60\%$). Notably, while TD3-Navigation and DWA-RL show fluctuating performance (e.g., TD3-Navigation's success rate drops to $65\%$ in scenario (d)), Ours demonstrates stable improvements, reducing collision rates by $15\textendash{}20\%$ compared to traditional methods in high-density settings.

Further analysis of navigation efficiency reveals that Ours balances path optimality and motion smoothness effectively. Although Potential Gap achieves the shortest average path length ($7.27$) in scenario (d), its success rate plummets to $60\%$, indicating compromised safety. In contrast, SafeMove achieves a moderate path length ($7.98$) while maintaining higher success rates, suggesting enhanced adaptability to complex scenarios. Additionally, SafeMove sustains stable average velocities ($0.16\textendash{}0.19$ m/s) across all environments, avoiding drastic speed reductions observed in TD3-Navigation ($0.15$ m/s in scenario (c)) or erratic velocity profiles in DWA-RL. This consistency underscores the algorithm's robustness in balancing speed and collision avoidance under escalating environmental complexity.

These results highlight Ours's scalability and reliability in dense dynamic environments. As obstacle counts rise from 8 to 20, Ours's success rate declines by only $20$ percentage points, markedly less than the $30\textendash{}35\%$ drops seen in DWA and TEB.

\subsection{Real-World Experiments}
The proposed algorithm was evaluated on a Spark two-wheel differential drive robot, which is equipped with an Intel RealSense D435 depth camera (with a field of view of 87° × 58° and a maximum detection range of 10 meters) and an onboard odometer for state information collection. To evaluate the static obstacle avoidance performance, experiments were conducted in two real environments: an indoor environment and a corridor (as shown in Fig. 7). The results demonstrated that the robot effectively avoided static obstacles and reached the target points in these scenarios. Additionally, the robot was tested in more complex environments, including static environments with dense obstacles and scenarios combining dynamic and static obstacles, as shown in Fig 7. The robot successfully navigated through these challenging environments and achieved the target points.
In summary, through real experiments conducted in four different environments, the robot achieved a success rate of over 70\% in dynamic obstacle avoidance planning. The results validate the effectiveness of the proposed algorithm in enhancing the robot's navigation capabilities in complex and dynamic environments.

\section{CONCLUSIONS}

In this study, we propose a novel dynamic obstacle avoidance strategy based on Deep Reinforcement Learning (DRL), enhancing the observation space and empirical replay buffer for better angular and linear velocities in dynamic environments. The algorithm is evaluated in both simulation and real-world settings, showing superior performance over state-of-the-art algorithms in collision rate, convergence speed, motion speed, and travel distance. However, it has limitations: (1) limited generalizability in real-world scenarios, requiring improvements in perception and reward functions; (2) potential performance degradation in more complex environments. Future work will focus on enhancing the algorithm’s robustness to optimize overall performance.

\addtolength{\textheight}{-12cm}   




\bibliographystyle{plain} 

\begin{thebibliography}{25}

\bibitem{c1} 
D. Fox, W. Burgard, and S. Thrun, “The dynamic window approach to collision avoidance,” \emph{IEEE Robot. Automat. Mag.}, vol. 4, no. 1, pp. 23–33, Mar. 1997, doi: 10.1109/100.580977.

\bibitem{c2} 
C. Rösmann, W. Feiten, T. Wösch, F. Hoffmann, and T. Bertram, “Trajectory modification considering dynamic constraints of autonomous robots”. https://doi.org/10.1007/s10514-012-9321-0.

\bibitem{c3} 
D. Wilkie, J. Van Den Berg, and D. Manocha, “Generalized velocity obstacles,” in \emph{2009 IEEE/RSJ International Conference on Intelligent Robots and Systems}, St. Louis, MO, USA: IEEE, Oct. 2009, pp. 5573–5578. doi: 10.1109/IROS.2009.5354175.

\bibitem{c4} 
E. Rimon and D. E. Koditschek, “Exact robot navigation using artificial potential functions,” \emph{IEEE Trans. Robot. Automat.}, vol. 8, no. 5, pp. 501–518, Oct. 1992, doi: 10.1109/70.163777.

\bibitem{c5} 
J. Minguez and L. Montano, “Nearness Diagram (ND) Navigation: Collision Avoidance in Troublesome Scenarios,” \emph{IEEE Trans. Robot. Automat.}, vol. 20, no. 1, pp. 45–59, Feb. 2004, doi: 10.1109/TRA.2003.820849.

\bibitem{c6} 
M. Mujahad, D. Fischer, B. Mertsching, and H. Jaddu, “Closest Gap based (CG) reactive obstacle avoidance Navigation for highly cluttered environments,” in \emph{2010 IEEE/RSJ International Conference on Intelligent Robots and Systems}, Taipei: IEEE, Oct. 2010, pp. 1805–1812. doi: 10.1109/IROS.2010.5649736.

\bibitem{c7} 
M. Mujahed and B. Mertsching, “The admissible gap (AG) method for reactive collision avoidance,” in \emph{2017 IEEE International Conference on Robotics and Automation (ICRA)}, Singapore: IEEE, May 2017, pp. 1916–1921. doi: 10.1109/ICRA.2017.8071093.

\bibitem{c8} 
R. Xu, S. Feng, and P. Vela, “Potential Gap: A Gap-Informed Reactive Policy for Safe Hierarchical Navigation,” \emph{IEEE Robot. Autom. Lett.}, vol. 6, no. 4, pp. 8325–8332, Oct. 2021, doi: 10.1109/LRA.2021.3104623.

\bibitem{c9} 
S. Feng, Z. Zhou, J. S. Smith, M. Asselmeier, Y. Zhao, and P. A. Vela, “GPF-BG: A Hierarchical Vision-Based Planning Framework for Safe Quadrupedal Navigation,” in \emph{2023 IEEE International Conference on Robotics and Automation (ICRA)}, London, United Kingdom: IEEE, May 2023, pp. 1968–1975. doi: 10.1109/ICRA48891.2023.10160804.

\bibitem{c10} 
H. Chen, S. Feng, Y. Zhao, C. Liu, and P. A. Vela, “Safe Hierarchical Navigation in Crowded Dynamic Uncertain Environments,” in \emph{2022 IEEE 61st Conference on Decision and Control (CDC)}, Cancun, Mexico: IEEE, Dec. 2022, pp. 1174–1181. doi: 10.1109/CDC51059.2022.9992674.

\bibitem{c11} 
R. Cimurs, I. H. Suh, and J. H. Lee, “Goal-Driven Autonomous Exploration Through Deep Reinforcement Learning,” \emph{IEEE Robot. Autom. Lett.}, vol. 7, no. 2, pp. 730–737, Apr. 2022, doi: 10.1109/LRA.2021.3133591.

\bibitem{c12} 
W. Huang, Y. Zhou, X. He, and C. Lv, “Goal-Guided Transformer Enabled Reinforcement Learning for Efficient Autonomous Navigation,” \emph{IEEE Trans. Intell. Transport. Syst.}, vol. 25, no. 2, pp. 1832–1845, Feb. 2024, doi: 10.1109/TITS.2023.3312453.

\bibitem{c13} 
S. Liu et al., “Intention Aware Robot Crowd Navigation with Attention-Based Interaction Graph,” in \emph{2023 IEEE International Conference on Robotics and Automation (ICRA)}, London, United Kingdom: IEEE, May 2023, pp. 12015–12021. doi: 10.1109/ICRA48891.2023.10160660.

\bibitem{c14} 
K. Bektas and H. I. Bozma, “APF-RL: Safe Mapless Navigation in Unknown Environments,” in \emph{2022 International Conference on Robotics and Automation (ICRA)}, Philadelphia, PA, USA: IEEE, May 2022, pp. 7299–7305. doi: 10.1109/ICRA46639.2022.9811537.

\bibitem{c15} 
Faust et al., “PRM-RL: Long-range Robotic Navigation Tasks by Combining Reinforcement Learning and Sampling-Based Planning,” in \emph{2018 IEEE International Conference on Robotics and Automation (ICRA)}, Brisbane, QLD: IEEE, May 2018, pp. 5113–5120. doi: 10.1109/ICRA.2018.8461096.

\bibitem{c16} 
H.-T. L. Chiang, J. Hsu, M. Fiser, L. Tapia, and A. Faust, “RL-RRT: Kinodynamic Motion Planning via Learning Reachability Estimators From RL Policies,” \emph{IEEE Robot. Autom. Lett.}, vol. 4, no. 4, pp. 4298–4305, Oct. 2019, doi: 10.1109/LRA.2019.2931199.

\bibitem{c17} 
H. Nguyen, S. H. Fyhn, P. De Petris, and K. Alexis, “Motion Primitives-based Navigation Planning using Deep Collision Prediction,” in \emph{2022 International Conference on Robotics and Automation (ICRA)}, Philadelphia, PA, USA: IEEE, May 2022, pp. 9660–9667. doi: 10.1109/ICRA46639.2022.9812231.

\bibitem{c18} 
U. Patel, N. K. S. Kumar, A. J. Sathyamoorthy, and D. Manocha, “DWA-RL: Dynamically Feasible Deep Reinforcement Learning Policy for Robot Navigation among Mobile Obstacles,” in \emph{2021 IEEE International Conference on Robotics and Automation (ICRA)}, Xi’an, China: IEEE, May 2021, pp. 6057–6063. doi: 10.1109/ICRA48506.2021.9561462.

\bibitem{c19} 
S. Fujimoto, H. van Hoof, and D. Meger, “Addressing Function Approximation Error in Actor-Critic Methods,” Oct. 22, 2018, arXiv: arXiv:1802.09477. doi: 10.48550/arXiv.1802.09477.

\bibitem{c20} 
J. Jia, X. Xing, and D. E. Chang, “GRU-Attention based TD3 Network for Mobile Robot Navigation,” in \emph{2022 22nd International Conference on Control, Automation and Systems (ICCAS)}, Jeju, Korea, Republic of: IEEE, Nov. 2022, pp. 1642–1647. doi: 10.23919/ICCAS55662.2022.10003950.

\bibitem{c21} 
B. Huang, J. Xie, and J. Yan, “Inspection Robot Navigation Based on Improved TD3 Algorithm,” \emph{Sensors}, vol. 24, no. 8, p. 2525, Apr. 2024, doi: 10.3390/s24082525.

\bibitem{c22} 
H. Liu, Y. Shen, C. Zhou, Y. Zou, Z. Gao, and Q. Wang, “TD3 Based Collision Free Motion Planning for Robot Navigation,” in \emph{2024 6th International Conference on Communications, Information System and Computer Engineering (CISCE)}, Guangzhou, China: IEEE, May 2024, pp. 247–250. doi: 10.1109/CISCE62493.2024.10653233.

\bibitem{c23} 
Bo Fu, Xulin Yao, "Research and Application of an Improved TD3 Algorithm in Mobile Robot Environment Perception and Autonomous Navigation", \emph{2024 3rd International Conference on Robotics, Artificial Intelligence and Intelligent Control (RAIIC)}, pp.158-162, 2024.

\bibitem{c24} 
T. Zhang, K. Zhang, J. Lin, W.-Y. G. Louie, and H. Huang, “Sim2real Learning of Obstacle Avoidance for Robotic Manipulators in Uncertain Environments,” \emph{IEEE Robot. Autom. Lett.}, vol. 7, no. 1, pp. 65–72, Jan. 2022, doi: 10.1109/LRA.2021.3116700.

\bibitem{c25} 
J. Van Den Berg, S. J. Guy, M. Lin, and D. Manocha, ``Reciprocal n-Body Collision Avoidance,'' in \emph{Robotics Research}, vol. 70, C. Pradalier, R. Siegwart, and G. Hirzinger, Eds., in \emph{Springer Tracts in Advanced Robotics}, vol. 70. , Berlin, Heidelberg: Springer Berlin Heidelberg, 2011, pp. 3--19. doi: 10.1007/978-3-642-19457-3\_1.

\end{thebibliography}

\end{document}